%% file: main.tex
\documentclass[10pt,twocolumn,letterpaper]{article}
 
\usepackage[pagenumbers]{cvpr} 


\definecolor{cvprblue}{rgb}{0.21,0.49,0.74}
\usepackage[pagebackref,breaklinks,colorlinks,allcolors=cvprblue]{hyperref}
\usepackage{bm}
\usepackage{multirow}
\usepackage{xcolor,colortbl}
\usepackage{amsmath}
\usepackage{wrapfig}
\usepackage{algorithm}
\usepackage{algcompatible}

\newtheorem{theorem}{Theorem}

\newcommand{\mymethod}{NCFM}
\newcommand{\DR}{\mathcal{D}}
\newcommand{\DS}{\tilde{\mathcal{D}}}
\definecolor{americanrose}{rgb}{1.0, 0.01, 0.24}
\def\paperID{5396} 
\def\confName{CVPR}
\def\confYear{2025}

\title{Dataset Distillation with Neural Characteristic Function: A Minmax Perspective}

\author{
    Shaobo Wang$^{1,2}$ \quad Yicun Yang$^{2}$ \quad  Zhiyuan Liu$^2$  \quad Chenghao Sun$^2$ \\
     \quad Xuming Hu$^3$ \quad Conghui He$^4$ \quad Linfeng Zhang$^{1,2}$\thanks{Corresponding Author.}
    \\ $^1$School of Artificial Intelligence, Shanghai Jiao Tong University \\
      $^2$EPIC Lab, Shanghai Jiao Tong University   \\
      $^3$Hong Kong University of Science and Technology, Guangzhou \\
      $^4$Shanghai Artificial Intelligence Laboratory
      \\  \texttt{\{shaobowang1009,zhanglinfeng\}@sjtu.edu.cn}
}

\begin{document}
\maketitle
\input{sec/0_abstract}
\input{sec/1_introduction}
\input{sec/2_related}
\input{sec/3_method}

\input{sec/4_experiment}
\input{sec/5_conclusion}
\clearpage
{
    \small
    \bibliographystyle{ieeenat_fullname}
    \bibliography{main}
}


\end{document}

%% file: sec/0_abstract.tex
\begin{abstract}
Dataset distillation has emerged as a powerful approach for reducing data requirements in deep learning. Among various methods, distribution matching-based approaches stand out for their balance of computational efficiency and strong performance. However, existing distance metrics used in distribution matching often fail to accurately capture distributional differences, leading to unreliable measures of discrepancy. In this paper, we reformulate dataset distillation as a minmax optimization problem and introduce Neural Characteristic Function Discrepancy (NCFD), a comprehensive and theoretically grounded metric for measuring distributional differences. NCFD leverages the Characteristic Function (CF) to encapsulate full distributional information, employing a neural network to optimize the sampling strategy for the CF's frequency arguments, thereby maximizing the discrepancy to enhance distance estimation. Simultaneously, we minimize the difference between real and synthetic data under this optimized NCFD measure. Our approach, termed Neural Characteristic Function Matching (\mymethod{}), inherently aligns the phase and amplitude of neural features in the complex plane for both real and synthetic data, achieving a balance between realism and diversity in synthetic samples. Experiments demonstrate that our method achieves significant performance gains over state-of-the-art methods on both low- and high-resolution datasets. Notably, we achieve a 20.5\% accuracy boost on ImageSquawk. Our method also reduces GPU memory usage by over 300$\times$ and achieves 20$\times$ faster processing speeds compared to state-of-the-art methods. To the best of our knowledge, this is the first work to achieve lossless compression of CIFAR-100 on a single NVIDIA 2080 Ti GPU using only 2.3 GB of memory. 
\end{abstract}

%% file: sec/1_introduction.tex
\section{Introduction}
\label{sec:intro}

Deep neural networks (DNNs) have achieved remarkable progress across a range of tasks, largely due to the availability of vast amounts of training data. However, training effectively with limited data remains challenging and crucial, particularly when large-scale datasets become too voluminous for storage. To address this, dataset distillation has been proposed to condense a large, real dataset into a smaller, synthetic one~\citep{DD,DC,DM,MTT,sre2l}. Dataset distillation has been applied in various areas, including neural architecture search~\citep{such2020generative,medvedev2021learning}, continual learning~\citep{An_Efficient_DC_Plugin_NeurIPS_2023,gu2024summarizing}, medical image computing~\citep{li2024image}, and privacy protection~\citep{chen2022private,chung2023rethinking,dong2022privacy}.

\begin{figure}[tb!]
    \centering
    \includegraphics[width=0.99 \linewidth]{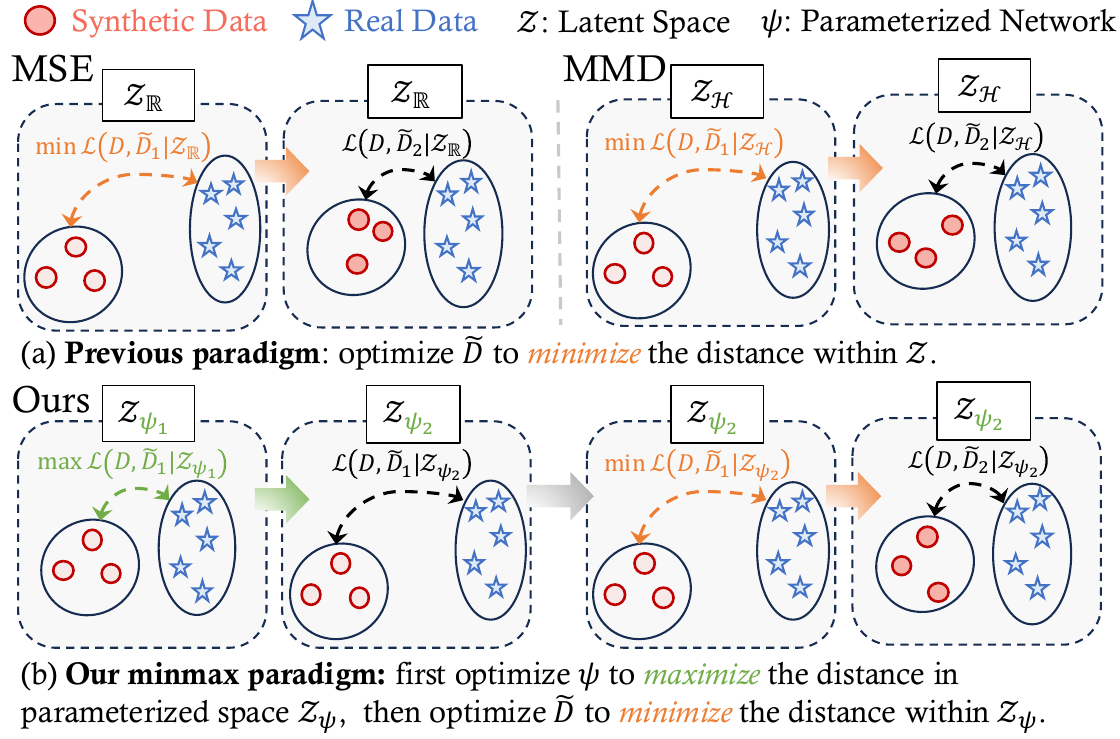}
    \vspace{-5pt}
    \caption{Comparison of different paradigms for dataset distillation. (a) The MSE approach compares point-wise features within Euclidean space, denoted as $ \mathcal{Z}_{\mathbb{R}} $, while MMD evaluates moment differences in Hilbert space, $ \mathcal{Z}_{\mathcal{H}} $. (b) Our method redefines distribution matching as a minmax optimization problem, where the distributional discrepancy is parameterized by a neural network $ \psi $. We begin by optimizing $ \psi $ to maximize the discrepancy, thereby establishing the latent space $ \mathcal{Z}_{\psi} $, and subsequently optimize the synthesized data $\DS$ to minimize this discrepancy within $\mathcal{Z}_{\psi}$.}
    \vspace{-10pt}
    \label{fig:metrics}
\end{figure}

\begin{figure*}[tb!]
    \centering
    \vspace{-5pt}
    \includegraphics[width=\linewidth]{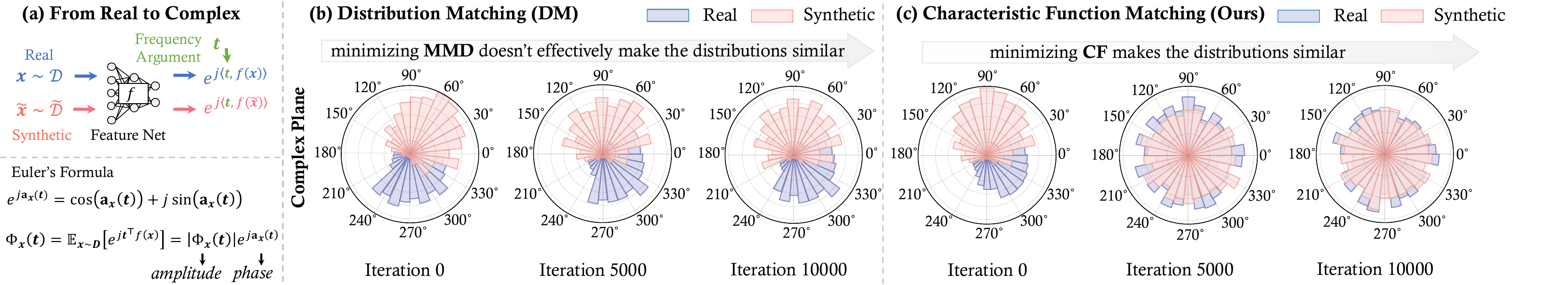}
    \vspace{-18pt}
    \caption{Comparison of different distribution matching methods. (a) Illustration of embedded features from the real domain to complex-plane features using Euler's formula~\citep{euler8transcending}. The latent neural feature $\Phi_{\bm{x}}(\bm{t})$ captures the amplitude and phase information. (b) MMD-based methods align feature moments in the embedded domain but may not effectively align the overall distributions. (c) CF-based methods directly compare distributions by balancing the amplitude and phase in the complex plane, enhancing distributional similarity.}
    \vspace{-5pt}
    \label{fig:MMD_vs_CF}
\end{figure*}

\begin{figure}[tb!]
    \vspace{-8pt}
    \centering
    \includegraphics[width=0.99\linewidth]{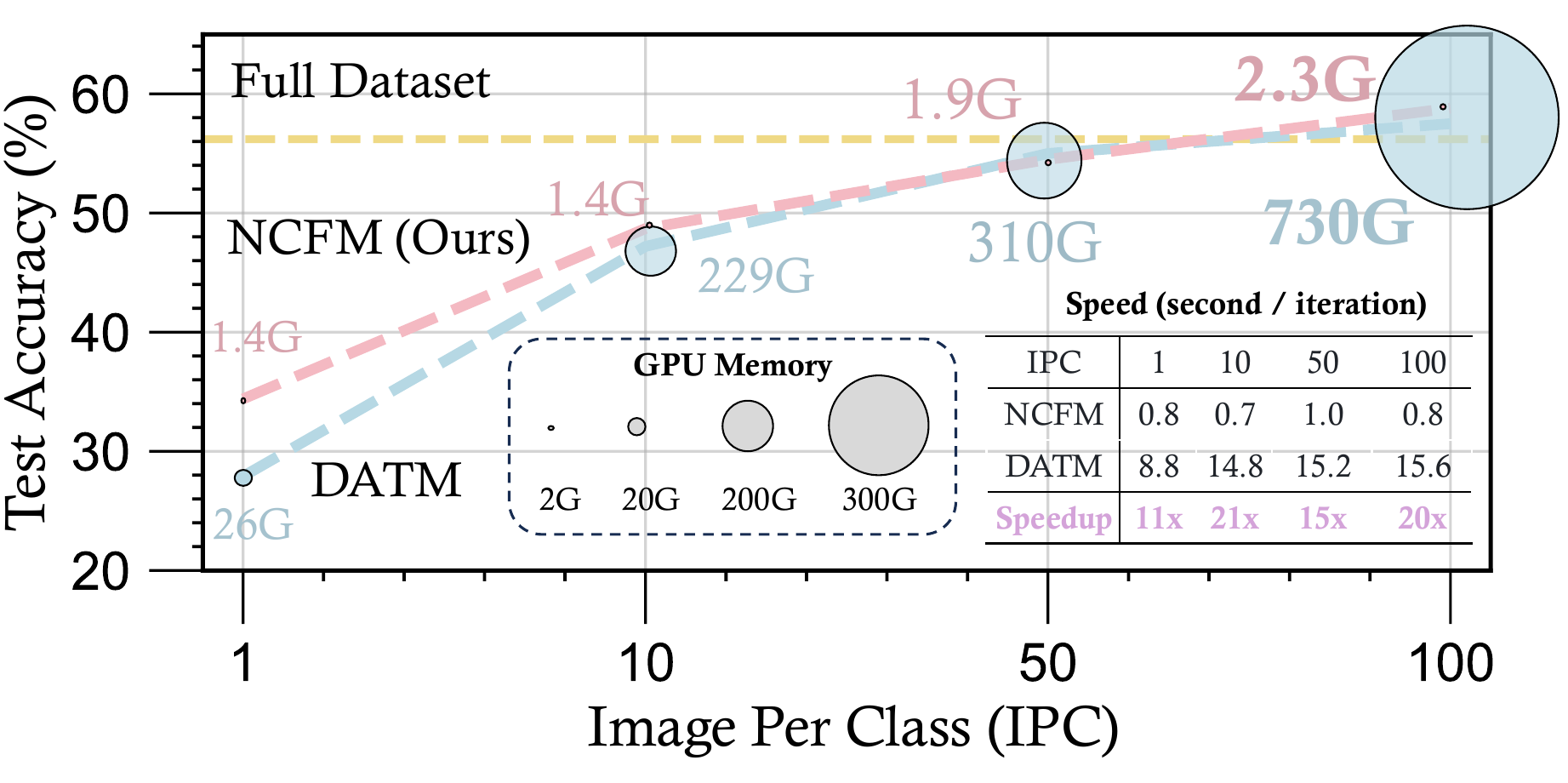}
    \vspace{-10pt}
    \caption{Comparison of performance, peak GPU memory usage, and distillation speed between the state-of-the-art (SOTA) distillation method and our \mymethod{} on CIFAR-100 across various IPC values, evaluated on 8 NVIDIA H100 GPUs. Notably, \mymethod{} reduces GPU memory usage by over 300$\times$, achieves 20$\times$ faster distillation, and delivers better performance. We also successfully demonstrated lossless  distillation using only 2.3GB GPU memory.}
    \label{fig:mem_speed}
    \vspace{-18pt}
\end{figure}

Among dataset distillation methods, feature or distribution matching (DM) approaches~\citep{CAFE,DM} have gained popularity for their effective balance between high performance and computational efficiency. Unlike bi-level optimization-based distillation approaches~\citep{DC,DSA,DCC,IDC,MTT}, DM-based methods bypass the need for nested optimization. For instance, when learning with 50 images per class (IPC) on CIFAR-10 dataset, DM methods achieve higher test accuracy than gradient matching methods~\citep{DC,DCC,DSA}, while requiring only a tenth of the computation time.

A key challenge in DM lies in defining an effective metric to measure distributional discrepancies between real and synthetic datasets. Early methods primarily employed Mean Squared Error (MSE) to compare point-wise features~\citep{CAFE,IID,DataDAM}, which operates in Euclidean space, $\mathcal{Z}_{\mathbb{R}}$, as illustrated on the left of Figure~\ref{fig:metrics}(a). However, MSE directly matches pixel-level or patch-level information without capturing the semantic structures embedded in high-dimensional manifolds, which falls short for distribution comparison. Later methods~\citep{DM,IDM,M3D} employ Maximum Mean Discrepancy (MMD) as a metric. Nevertheless, research in generative modeling~\citep{li2017mmd,binkowski2018demystifying} has shown that MMD aligns moments of distributions in a latent Hilbert space, $\mathcal{Z}_{\mathcal{H}}$, as shown on the right of Figure~\ref{fig:metrics}(a). While distributional equivalence implies moment equivalence, the converse is not necessarily true: aligning moments alone does not guarantee full distributional matching. As illustrated in Figure~\ref{fig:MMD_vs_CF}(b), MMD-based methods may fail to capture overall distributional alignment between real and synthetic data, resulting in suboptimal synthesized image quality.

To overcome these limitations, we propose a novel approach that reformulates distribution matching as an \textit{adversarial minmax optimization problem}, as depicted in Figure~\ref{fig:metrics}(b). By leveraging the minmax paradigm, we adaptively learn the discrepancy metric, enabling it to \textit{maximize} the separability between real and synthetic data distributions. This dynamic adjustment addresses the rigidity of fixed metrics like MSE and MMD. Meanwhile, the synthetic data is iteratively optimized to \textit{minimize} the dynamically refined discrepancy measure. Building upon this foundation, we introduce Neural Characteristic Discrepancy (NCFD), a parameterized metric based on the Characteristic Function (CF), which provides a precise and comprehensive representation of the underlying probability distribution. Defined as the Fourier transform of the probability density function, the CF encapsulates all relevant information about a distribution~\citep{levy1937,kogon1998characteristic,billingsley2017probability,bisgaard2000characteristic,feuerverger1977empirical,shephard1991characteristic}. The CF offers a one-to-one correspondence with the cumulative density function, ensuring the robustness and reliability.

In our framework, an auxiliary network embeds features while a lightweight sampling network is optimized to dynamically adjust its CF sampling strategy using a scale mixture of normals. During the distillation process, we iteratively minimize the NCFD to bring synthetic data closer to real data, while training the sampling network to maximize NCFD, thereby improving the metric’s robustness and accuracy. Unlike MMD which has quadratic computational complexity,  NCFD achieves linear time computational complexity. Our method, Neural Characteristic Function Matching (\mymethod{}), aligns both the phase and amplitude of neural features in the complex plane, achieving a balanced synthesis of realism and diversity in the generated images. As shown in Figure~\ref{fig:MMD_vs_CF}(c), \mymethod{} effectively captures overall distributional information, leading to well-aligned synthetic and real data distributions after optimization. Our contributions are as follows:
\begin{enumerate}[leftmargin=10pt, topsep=0pt, itemsep=1pt, partopsep=1pt, parsep=1pt]
\item We reformulate the distribution matching problem as a minmax optimization problem, where the sampling network maximizes the distributional discrepancy to learn a proper discrepancy metric, while the synthesized images are optimized to minimize such discrepancy.
\item We introduce Neural Characteristic Function Matching (\mymethod{}), which aligns the phase and amplitude information of neural features in the complex plane for both real and synthetic data, achieving a balance between realism and diversity in synthetic data.
\item Extensive experiments across multiple benchmark datasets demonstrate the superior performance and efficiency of \mymethod{}. Particularly, on high-resolution datasets, \mymethod{} achieves significant accuracy gains of up to 20.5\% on ImageSquawk and 17.8\% on ImageMeow at 10 IPC compared to SOTA methods.
\item \mymethod{} achieves unprecedented efficiency in computational resources. As shown in Figure~\ref{fig:mem_speed}, our method dramatically reduces resource requirements with better performance, \textbf{achieving more than 300$\times$ reduction in GPU memory usage compared with DATM~\citep{DATM}}. Most remarkably, \textbf{\mymethod{} demonstrates lossless dataset distillation on both CIFAR-10 and CIFAR-100 using about merely 2GB GPU memory}, enabling all experiments to be conducted on a single NVIDIA 2080 Ti GPU. 
\end{enumerate}

%% file: sec/2_related.tex
\vspace{-5pt}
\section{Related Work}
\label{sec:related}

\noindent \textbf{Dataset Distillation Methods Based on Distribution and Feature Matching.} Dataset distillation was proposed by~\citep{DD}. Compared with various bi-level DD methods, DM~\citep{DM} is regarded as a efficient method that balances the performance and computational efficiency, without involving the nested model optimization. These methods can be classified into two directions, \emph{i.e.}, point-wise and moment-wise matching. For moment-wise matching, DM-based methods~\citep{DM,IDM,M3D} propose to minimize the maximum mean discrepancy (MMD) between synthetic and real datasets. For point-wise feature matching, they typically design better strategies to match features extracted across layers in convolutional neural networks, and apply further adjustments to improve the performance~\citep{CAFE,IID,DataDAM}. However, moment-based and point-based matching methods may not capture the overall distributional discrepancy between synthetic and real data, as they are not sufficient conditions for distributional equivalence.

\begin{figure*}[tb!]
    \centering
    \vspace{-2em}
    \captionsetup{type=figure}
    \includegraphics[width=0.9\linewidth]{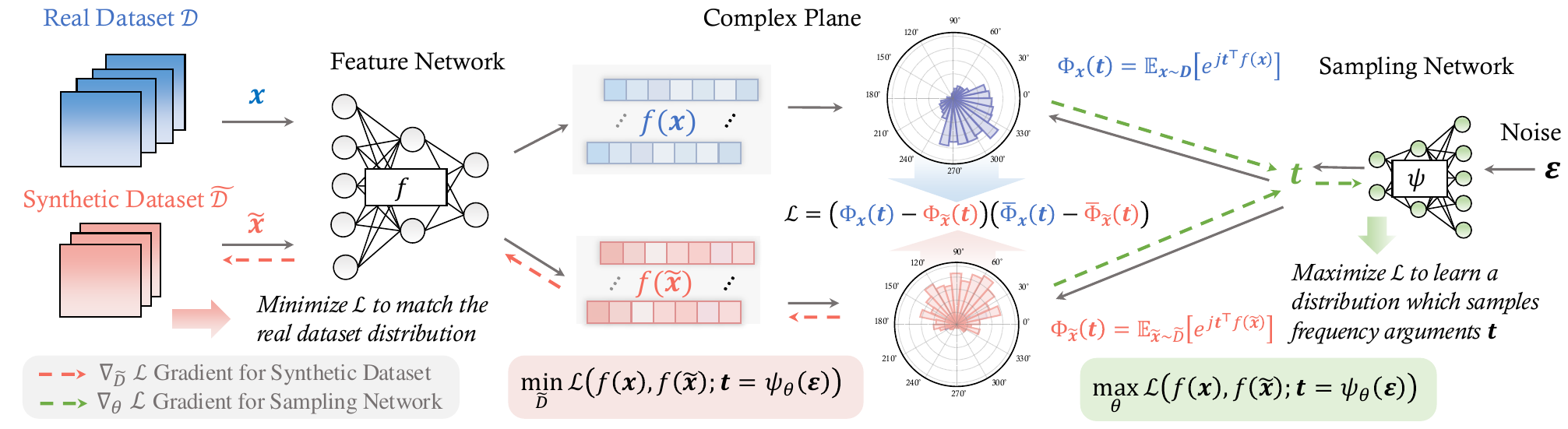}
    \vspace{-5pt}
    \captionof{figure}{Dataset Distillation with Neural Characteristic Function Matching (\mymethod{}). Real and synthetic data points are sampled and embedded through a feature extractor network. The synthetic data is optimized by minimizing the distributional discrepancy between real and synthetic data, measured via the Neural Characteristic Function Discrepancy (NCFD) in the complex plane. Additionally, an auxiliary network learns an optimal sampling distribution for the frequency arguments of the characteristic function. Best viewed in color.}
    \vspace{-5pt}
    \label{fig:NCFM_pipeline}
\end{figure*}%
\noindent \textbf{Characteristic Function as a Distributional Metric.} The characteristic function is a unique and universal metric for measuring distributional discrepancy, defined as the Fourier transform of the probability density function~\citep{billingsley2017probability}. The CF of any real-valued random variable completely defines its probability distribution, providing an alternative analytical approach compared to working directly with probability density or cumulative distribution functions. Unlike the moment-generating function, the CF always exists when treated as a function of a real-valued argument~\citep{bisgaard2000characteristic}. Recently, the CFD has been adopted in deep learning for various tasks, \emph{e.g.}, several works have been proposed to use the CFD for generative modeling~\citep{CFGAN,RCFGAN}. However, none of prior works have considered the CFD for  distillation.

%% file: sec/3_method.tex
\section{Preliminaries: Distribution Matching}
Distribution Matching (DM) was first introduced by~\citep{DM} as an alternative to traditional bi-level optimization techniques, such as gradient matching methods~\citep{DC,IDC,DSA,DCC} and trajectory matching methods~\citep{MTT,TESLA,FTD,DATM}. Classical DM approaches focus on minimizing the discrepancy between the distributions of real and synthetic data, typically categorized into two main types: feature point matching and moment matching. Feature point matching methods~\citep{CAFE,IID,DataDAM} directly compare point-wise features using Mean Square Error (MSE), as defined by:
\begin{equation}
\vspace{-5pt}
    \mathcal{L}_{\textrm{MSE}} = \mathbb{E}_{\bm{x} \sim \DR, \bm{\tilde{x}} \sim \DS} \left[\left\|f(\bm{x}) - f(\tilde{\bm{x}})\right\|^2\right],
\end{equation}
where $f$ denotes the feature extractor network, $\DR$ and $\DS$ represent the real and synthetic data distributions, respectively, $\bm{x}$ and $\bm{\tilde{x}}$ are samples drawn from $\DR$ and $\DS$. However, MSE may not be ideal for comparing distributions, as it only considers direct feature comparisons in Euclidean space, neglecting important semantic information.

In another line, notable works employed Maximum Mean Discrepancy (MMD) to align high-order moments in the latent feature space~\citep{DM,IDM,M3D}. Rigorously, MMD is defined to match moments within the Reproducing Kernel Hilbert Space (RKHS) induced by a selected kernel function. The MMD loss can be formulated as:
\begin{equation}
    \begin{aligned}
    & \sup_{f \in \mathcal{F}} \left\|\mathbb{E}_{\bm{x} \sim \DR} \left[f(\bm{x})\right] - \mathbb{E}_{\bm{\tilde{x}} \sim \DS} \left[f(\tilde{\bm{x}})\right]\right\|^2, \\
    = &  \sup_{f \in \mathcal{F}} \left(\mathcal{K}_{\DR,\DR} + \mathcal{K}_{\DS,\DS} - 2\mathcal{K}_{\DR,\DS}\right),
    \end{aligned}
\end{equation}
where $\mathcal{K}_{\DR,\DS}=\mathbb{E}_{\bm{x}\sim \DR, \tilde{\bm{x}} \sim \DS}[\mathcal{K}_{f(\bm{x}),f(\tilde{\bm{x}})}]$ denotes the kernel function associated with feature extractor $f$ in function class $\mathcal{F}$. The choice of kernel function $\mathcal{K}$ is not unique and requires careful selection for MMD to be effective. However, instead of selecting certain kernel function, most DM-based methods~\citep{DM,IDM,IID} align moments directly in the feature space, commonly approximated as:
\begin{equation}
\vspace{-3pt}
    \mathcal{L}_{\textrm{MMD}} = \left\|\mathbb{E}_{\bm{x} \sim \DR} \left[f(\bm{x})\right] - \mathbb{E}_{\bm{\tilde{x}} \sim \DS} \left[f(\tilde{\bm{x}})\right]\right\|^2.
\end{equation}
We argue that such empirical MMD estimates lack rigor, as they do not provide a maximal upper bound on the discrepancy, falling short of MMD’s theoretical requirements.

\section{Adversarial Distribution Matching}
\subsection{Minmax Framework}
To address existing challenges with discrepancy measure selection, we propose a new approach that reformulates distribution matching as a minmax optimization problem. In this framework, we maximize the discrepancy measure to define a robust discrepancy metric, parameterized by a neural network $\psi$. Concurrently, we minimize the discrepancy between the synthetic dataset $\DS$ and the real dataset $\DR$ by optimizing the synthetic data distribution $\DS$. Formally, this minmax optimization problem is expressed as:
\begin{equation}
\vspace{-5pt}
    \min_{\DS} \max_{\psi} \mathcal{L}(\DS, \DR, f, \psi),
\end{equation}
where $\mathcal{L}$ denotes the discrepancy measure, $f$ is the feature extractor network, and $\psi$ is the network learning the discrepancy metric. This minmax framework seeks the optimal synthetic data distribution $\DS$ that minimizes $\mathcal{L}$ while network $\psi$ maximizes $\mathcal{L}$ to establish a robust  metric. 

\subsection{Neural Characteristic Function Matching}
\subsubsection{Neural Characteristic Function Discrepancy}
To define a suitable discrepancy metric within this minmax framework, we propose a novel discrepancy measure based on the characteristic function (CF), which enables direct and robust assessment of distributional discrepancies. Characteristic functions are a mainstay in probability theory, often used as an alternative to probability density functions due to their unique properties. Specifically, the CF of a random variable $\bm{x}$ is the expectation of its complex exponential transform, defined as:
\begin{equation}\label{eq:CF}
\vspace{-5pt}
     \Phi_{\bm{x}}(\bm{t})=\mathbb{E}_{\bm{x}}\left[e^{j \langle \bm{t},\bm{x} \rangle }\right] = \int_{\bm{x}} e^{j \langle \bm{t},\bm{x} \rangle } dF(\bm{x}),
\end{equation}
where $F(\bm{x})$ denotes the cumulative distribution function (cdf) of $\bm{x}$, $j=\sqrt{-1}$, and $\bm{t}$ is the frequency argument. Since the cdf is not directly accessible in practice, we approximate the CF empirically as $\Phi_{\bm{x}}(\bm{t}) = \frac{1}{N} \sum_{i=1}^{N} e^{j \langle \bm{t},\bm{x}_i \rangle }$, where $N$ is the sample size in the dataset. The CF provides a theoretically principled description of a distribution, summarized in the following theorems.
\begin{theorem}[Lévy's Convergence Theorem~\citep{levy1937}]\label{thm:convergence}
    Let $\{X_n\}_{n=1}^{\infty}$ be a sequence of random variables with characteristic functions $\Phi_n(\bm{t}) = \mathbb{E}_{X_n}\left[e^{j \langle \bm{t}, X_n \rangle }\right]$. Suppose $\Phi_n(\bm{t}) \to \Phi(\bm{t})$ pointwise for each $\bm{t} \in \mathbb{R}^d$ as $n \to \infty$. If $\Phi(\bm{t})$ is continuous at $\bm{t} = 0$, then there exists a random variable $X$ with characteristic function $\Phi(\bm{t})$, and $X_n$ converges in distribution to $X$.
\end{theorem}
\begin{theorem}[Uniqueness for Characteristic Functions~\citep{feuerverger1977empirical}]\label{thm:unique}
\vspace{-8pt}
    If two random variables $X$ and $Y$ have the same characteristic function, $\Phi_X(\bm{t}) = \Phi_Y(\bm{t})$ for all $\bm{t}$, then $X$ and $Y$ are identically distributed. In other words, a characteristic function uniquely determines the distribution.
\end{theorem}
By Theorems~\ref{thm:convergence} and~\ref{thm:unique}, the empirical CF weakly converges to the population CF, ensuring that the CF serves as a reliable proxy for the underlying distribution. Based on the CF, we define our characteristic function discrepancy (CFD) as:
\begin{equation}\label{eq:chfloss}
    \mathcal{C}_{\mathcal{T}}(\bm{x}, \bm{\tilde{x}}) \!=\!\int_{\bm{t}} \sqrt{\underbrace{({\Phi}_{\bm{x}}(\bm{t}) \!-\! {\Phi}_{\bm{\tilde{x}}}(\bm{t}))(\bar{\Phi}_{\bm{x}}(\bm{t}) \!-\! \bar{\Phi}_{\bm{\tilde{x}}}(\bm{t}))}_{\textrm{Chf}(\bm{t})}} dF_{\mathcal{T}}(\bm{t}),
\end{equation}
where $\bar{\Phi}$ is the complex conjugate of $\Phi$, and $F_{\mathcal{T}}(\bm{t})$ is the CDF of a sampling distribution on $\bm{t}$. To simplify, we let $\textrm{Chf}(\bm{t}) = (\Phi_{\bm{x}}(\bm{t}) - \Phi_{\bm{\tilde{x}}}(\bm{t}))(\bar{\Phi}_{\bm{x}}(\bm{t}) - \bar{\Phi}_{\bm{\tilde{x}}}(\bm{t}))$ for further analysis. We show that $\mathcal{C}_{\mathcal{T}}(\bm{x}, \bm{\tilde{x}})$ defines a valid distance metric in the following theorem.
\begin{theorem}[CFD as a Distance Metric.]\label{chfloss_distance}
    The CF discrepancy $\mathcal{C}_{\mathcal{T}}(\bm{x}, \bm{\tilde{x}})$, as defined in Eq.~(\ref{eq:chfloss}), serves as a distance metric between $\bm{x}$ and $\bm{\tilde{x}}$ when the support of $\mathcal{T}$ resides in Euclidean space. It satisfies the properties of \textit{non-negativity}, \textit{symmetry}, and the \textit{triangle inequality}.
\end{theorem}
Theorem~\ref{chfloss_distance} establishes that CFD is a valid distance metric. Furthermore, we demonstrate that this formulation decomposes CFD into phase, $\bm{a}_{\bm{x}}(\bm{t})$, and amplitude, $|{\Phi}_{\bm{x}}(\bm{t})|$, components through Euler’s formula:
\begin{equation}
\vspace{-5pt}
    \begin{aligned}
        & \textrm{Chf}(\bm{t})   = \left|{\Phi}_{\bm{x}}(\bm{t})\right|^2 + \left|{\Phi}_{\bm{\tilde{x}}}(\bm{t})\right|^2 \\ 
        & - \left|{\Phi}_{\bm{x}}(\bm{t})\right|\left|{\Phi}_{\bm{\tilde{x}}}(\bm{t})\right| (2 \cos(\bm{a}_{\bm{x}}(\bm{t}) - \bm{a}_{\bm{\tilde{x}}}(\bm{t}))) \\
        & = \underbrace{\left(\left|{\Phi}_{\bm{x}}(\bm{t})-{\Phi}_{\bm{\tilde{x}}}(\bm{t})\right|\right)^2}_{\text{amplitude difference}} \\ 
        & + 2 \left|{\Phi}_{\bm{x}}(\bm{t})\right|\left|{\Phi}_{\bm{\tilde{x}}}(\bm{t})\right| \underbrace{(1 - \cos(\bm{a}_{\bm{x}}(\bm{t}) - \bm{a}_{\bm{\tilde{x}}}(\bm{t})))}_{\text{phase difference}},
    \end{aligned}
\end{equation}

\noindent $\bullet$ \textit{Phase Information:} the term $1 - \cos(\bm{a}_{\bm{x}}(\bm{t}) - \bm{a}_{\bm{\tilde{x}}}(\bm{t}))$ represents the phase, encoding data centres crucial for \textit{realism}.

\noindent $\bullet$ \textit{Amplitude Information:} the term $\left|\Phi_{\bm{x}}(\bm{t}) - \Phi_{\bm{\tilde{x}}}(\bm{t})\right|^2$ captures the distribution scale, contributing to the \textit{diversity}.

This phase-amplitude decomposition, supported in signal processing~\citep{oppenheim1981importance,mandic2009complex} and generative models~\citep{RCFGAN}, provides insight into CFD's descriptive power. In practice, to reduce computational cost, we furehr introduce an additional feature extractor  $f$  to map input variables into a latent space, which is similar to previous works in distribution matching~\citep{DM,IDM,IID,DSDM}. We also use a parameterized sampling network $\psi$ to obtain the distribution of frequency argument $\bm{t}$, thereby extending the CFD to a more general parameterized form, \emph{i.e.}, \emph{Neural Characteristic Function Discrepancy (NCFD)} as $   \mathcal{C}_{\mathcal{T}}(\bm{x}, \bm{\tilde{x}}; f,\psi) = \int_{\bm{t}} \sqrt{\textrm{Chf}(\bm{t};f)} dF_{\mathcal{T}}(\bm{t};\psi)$, where $\textrm{Chf}(\bm{t};f)$ is  defined as $\left(\left|{\Phi}_{f(\bm{x})}(\bm{t})-{\Phi}_{f(\bm{\tilde{x}})}(\bm{t})\right|\right)^2 + 2 \left|{\Phi}_{f(\bm{x})}(\bm{t})\right|\left|{\Phi}_{f(\bm{\tilde{x}})}(\bm{t})\right| (1 - \cos(\bm{a}_{f(\bm{x})}(\bm{t}) - \bm{a}_{f(\bm{\tilde{x}})}(\bm{t})))$.

\subsubsection{Determining the sampling strategy for NCFD}
The core aspect in optimizing $\mathcal{C}_{\mathcal{T}}(\bm{x}, \bm{\tilde{x}}; f, \psi)$ lies in determining the form of $F_{\mathcal{T}}(\bm{t};\psi)$, \emph{i.e.}, how to correctly and efficiently sample $\bm{t}$ from a carefully picked distribution. Similar with works in generative adversarial network~\citep{CFGAN,NCFGAN}, we define $F_{\mathcal{T}}(\bm{t})$ as the cumulative distribution function (cdf) of a \textit{scale mixture of normals}, as $    p_{\mathcal{T}}(\bm{t}) = \int_{\bm{\Sigma}} \bm{\mathcal{N}}(\bm{t}|\bm{0}, \bm{\Sigma})p_{\bm{\Sigma}}(\bm{\Sigma})d\bm{\Sigma}$, where $p_{\mathcal{T}}(\bm{t})$ is the probability density function (pdf) of $F_{\mathcal{T}}(\bm{t})$, $\bm{\mathcal{N}}(\bm{t}|\bm{0}, \bm{\Sigma})$ denotes a zero-mean Gaussian distribution with covariance $\bm{\Sigma}$, and $p_{\bm{\Sigma}}(\bm{\Sigma})$ represents the distribution of $\bm{\Sigma}$. We observe that as the number of sampled frequency arguments increases, the approximation of the empirical CF improves, as guaranteed by Lévy's Convergence Theorem~\citep{levy1937}, ultimately leading to higher quality synthetic data. 

\subsubsection{Distribution Matching with NCFD}
Given the NCFD measure $\mathcal{C}_{\mathcal{T}}(\bm{x}, \bm{\tilde{x}}; f, \psi)$, we now propose a method to utilize NCFD for distribution matching, termed as Neural Characteristic Function Matching (NCFM). A visual illustration of the NCFM pipeline is provided in Figure~\ref{fig:NCFM_pipeline}. On one hand, we maximize the NCFD to learn an effective discrepancy metric by optimizing the network $\psi$. On the other hand, we minimize this learned NCFD to obtain an optimal synthetic dataset, $\DS$. In practice, we introduce a hyper-parameter $\alpha$ to balance the amplitude and phase information in the NCFD, then the minmax optimization problem can be formulated as:
\begin{equation}\label{eq:alpha_blend}
\begin{small}
    \begin{aligned}
        & \min_{\DS} \max_{\psi} \mathcal{L}(\DS, \DR, f, \psi) = \min_{\DS} \max_{\psi} \mathbb{E}_{\bm{x}\sim \DR, \bm{\tilde{x}} \sim \DS} \mathcal{C}_{\mathcal{T}}(\bm{x}, \bm{\tilde{x}}; f, \psi) \\
        & =  \min_{\DS} \max_{\psi} \mathbb{E}_{\bm{x}\sim \DR, \bm{\tilde{x}} \sim \DS} \int_{\bm{t}} \sqrt{\textrm{Chf}(\bm{t};f)} \  dF_{\mathcal{T}}(\bm{t};\psi) \\
      & \text{where} \quad \textrm{Chf}(\bm{t};f) = \alpha \left(\left(\left|{\Phi}_{f(\bm{x})}(\bm{t})-{\Phi}_{f(\bm{\tilde{x}})}(\bm{t})\right|\right)^2\right) + (1-\alpha) \cdot \\
       &  (2 \left|{\Phi}_{f(\bm{x})}(\bm{t})\right|\left|{\Phi}_{f(\bm{\tilde{x}})}(\bm{t})\right|) \cdot (1 - \cos(\bm{a}_{f(\bm{x})}(\bm{t}) - \bm{a}_{f(\bm{\tilde{x}})}(\bm{t}))).
    \end{aligned}
\end{small}
\end{equation}
For the design of $ f $, we used a hybrid approach that combines a pre-trained model with a randomly initialized model, both selected from a subset of trained models. This ensures that the feature extractor remains moderately diverse yet discriminative. The hybrid feature extractor is constructed by $\beta$-blending the checkpoints of the initial and final models, where each model is chosen from a specific subset of available models. At each distillation step, the blending coefficient $\beta \in (0,1)$ is sampled from a uniform distribution $\mathcal{U}(0,1)$, providing a balanced combination of initial and final checkpoints. Our \mymethod{} can be seamlessly integrated with additional data curation steps, such as generating soft labels with a pre-trained neural network and performing dataset finetuning. Unlike prior methods that focus on learning soft labels~\citep{ladd,nodistill,DATM}, \mymethod{} simply leverages a pre-trained network to efficiently generate soft labels for the distilled dataset, improving both efficiency and effectiveness. However, these additional curation steps are not essential for \mymethod{}, as it can achieve SOTA performance within the pure minmax framework.

%% file: sec/4_experiment.tex
\begin{table*}[tb!]
    \centering
     \vspace{-4pt}
    \caption{Results of \mymethod{} on CIFAR-10/100, and Tiny ImageNet (resolution of 64×64) datasets.}
    \label{tab:dd_main_exp}
    \vspace{-8pt}
    \setlength{\tabcolsep}{4mm}{\resizebox{0.95\textwidth}{!}{
    \begin{tabular}{c|ccc|ccc|ccc}
    \toprule
     Dataset & \multicolumn{3}{c|}{CIFAR-10}& \multicolumn{3}{c|}{CIFAR-100}& \multicolumn{3}{c}{Tiny ImageNet} \\ 
    IPC & 1 & 10 & 50 & 1 & 10 & 50 & 1 & 10 & 50 \\ 
    Ratio (\%) & 0.02 & 0.2 & 1 & 0.2 & 2 & 10 & 0.2 & 2  & 10\\\midrule
    Random & 14.4{\scriptsize$\pm$2.0} & 26.0{\scriptsize$\pm$1.2} & 43.4{\scriptsize$\pm$1.0} & 4.2{\scriptsize$\pm$0.3} & 14.6{\scriptsize$\pm$0.5} & 30.0{\scriptsize$\pm$0.4} & 1.4{\scriptsize$\pm$0.1} & 5.0{\scriptsize$\pm$0.2} & 15.0{\scriptsize$\pm$0.4} \\ 
    Herding & 21.5{\scriptsize$\pm$1.2} & 31.6{\scriptsize$\pm$0.7} & 40.4{\scriptsize$\pm$0.6} & 8.4{\scriptsize$\pm$0.3} & 17.3{\scriptsize$\pm$0.3} & 33.7{\scriptsize$\pm$0.5} & 2.8{\scriptsize$\pm$0.2} & 6.3{\scriptsize$\pm$0.2} & 16.7{\scriptsize$\pm$0.3} \\ 
    Forgetting & 13.5{\scriptsize$\pm$1.2} & 23.3{\scriptsize$\pm$1.0} & 23.3{\scriptsize$\pm$1.1} & 4.5{\scriptsize$\pm$0.2} & 15.1{\scriptsize$\pm$0.3} & 30.5{\scriptsize$\pm$0.3} & 1.6{\scriptsize$\pm$0.1} & 5.1{\scriptsize$\pm$0.2} & 15.0{\scriptsize$\pm$0.3} \\ 
    \midrule
    DC & 28.3{\scriptsize$\pm$0.5 } & 44.9{\scriptsize$\pm$0.5 } & 53.9{\scriptsize$\pm$0.5 } & 12.8{\scriptsize$\pm$0.3 } & 25.2{\scriptsize$\pm$0.3 } & - & -& - & -\\ 
    DSA & 28.8{\scriptsize$\pm$0.7 } & 52.1{\scriptsize$\pm$0.5 } & 60.6{\scriptsize$\pm$0.5 } &  13.9{\scriptsize$\pm$0.3 } & 32.3{\scriptsize$\pm$0.3 } & 42.8{\scriptsize$\pm$0.4 } & - & -& -\\ 
    DCC & 32.9{\scriptsize$\pm$0.8 } & 49.4{\scriptsize$\pm$0.5 } & 61.6{\scriptsize$\pm$0.4 } & 13.3{\scriptsize$\pm$0.3 } & 30.6{\scriptsize$\pm$0.4 } & 40.0{\scriptsize$\pm$0.3 } & -& -& - \\ 
    DSAC & 34.0{\scriptsize$\pm$0.7 } & 54.5{\scriptsize$\pm$0.5 } & 64.2{\scriptsize$\pm$0.4 } & 14.6{\scriptsize$\pm$0.3 } & 14.6{\scriptsize$\pm$0.3 } & 39.3{\scriptsize$\pm$0.4 } & -& -& - \\ 
    FrePo & 46.8{\scriptsize$\pm$0.7 } & 65.5{\scriptsize$\pm$0.4 } & 71.7{\scriptsize$\pm$0.2 } &  28.7{\scriptsize$\pm$0.1 } & 42.5{\scriptsize$\pm$0.2 } & 44.3{\scriptsize$\pm$0.2 } &15.4{\scriptsize$\pm$0.3 } & 25.4{\scriptsize$\pm$0.2 } & -\\
    MTT & 46.3{\scriptsize$\pm$0.8 } & 65.3{\scriptsize$\pm$0.7 } & 71.6{\scriptsize$\pm$0.2 } & 24.3{\scriptsize$\pm$0.3 } & 40.1{\scriptsize$\pm$0.4 } & 47.7{\scriptsize$\pm$0.2 } &  8.8{\scriptsize$\pm$0.3 } & 23.2{\scriptsize$\pm$0.2 } & 28.0{\scriptsize$\pm$0.3} \\
    ATT& 48.3{\scriptsize$\pm$1.0 } & 67.7{\scriptsize$\pm$0.6 } & 74.5{\scriptsize$\pm$0.4 } &  26.1{\scriptsize$\pm$0.3 } & 44.2{\scriptsize$\pm$0.5 } & 51.2{\scriptsize$\pm$0.3 } & 11.0{\scriptsize$\pm$0.5 } & 25.8{\scriptsize$\pm$0.4 } & - \\
    FTD & 46.8{\scriptsize$\pm$0.3 } & 66.6{\scriptsize$\pm$0.3 } & 73.8{\scriptsize$\pm$0.2 } &  25.2{\scriptsize$\pm$0.2 } & 43.4{\scriptsize$\pm$0.3 } & 48.5{\scriptsize$\pm$0.3 } & 10.4{\scriptsize$\pm$0.3 } & 24.5{\scriptsize$\pm$0.2 } & - \\
    TESLA & 48.5{\scriptsize$\pm$0.8 } & 66.4{\scriptsize$\pm$0.8 } & 72.6{\scriptsize$\pm$0.7 } & 24.8{\scriptsize$\pm$0.4 } & 41.7{\scriptsize$\pm$0.3 } & 47.9{\scriptsize$\pm$0.3 } & -  & -  & - \\
    CAFE & 30.3{\scriptsize$\pm$1.1 } & 46.3{\scriptsize$\pm$0.6 } & 55.5{\scriptsize$\pm$0.6 } &  12.9{\scriptsize$\pm$0.3 } & 27.8{\scriptsize$\pm$0.3 } & 37.9{\scriptsize$\pm$0.3 } & - & - & - \\ 
    DM & 26.0{\scriptsize$\pm$0.8 } & 48.9{\scriptsize$\pm$0.6 } & 63.0{\scriptsize$\pm$0.4 } &  11.4{\scriptsize$\pm$0.3 } & 29.7{\scriptsize$\pm$0.3 } & 43.6{\scriptsize$\pm$0.4 } & 3.9{\scriptsize$\pm$0.2 } & 12.9{\scriptsize$\pm$0.4 } & 24.1{\scriptsize$\pm$0.3} \\
    IDM & 45.6{\scriptsize$\pm$0.7 } & 58.6{\scriptsize$\pm$0.1 } & 67.5{\scriptsize$\pm$0.1 } & 20.1{\scriptsize$\pm$0.3 } & 45.1{\scriptsize$\pm$0.1 } & {50.0{\scriptsize$\pm$0.2 }} & 10.1{\scriptsize$\pm$0.2 } & 21.9{\scriptsize$\pm$0.2 } & 27.7{\scriptsize$\pm$0.3 } \\ 
    M3D & 45.3{\scriptsize$\pm$0.3 } & 63.5{\scriptsize$\pm$0.2 } & 69.9{\scriptsize$\pm$0.5 } & 26.2{\scriptsize$\pm$0.3 } & 42.4{\scriptsize$\pm$0.2 } & {50.9{\scriptsize$\pm$0.7 }} & - & - & - \\ 
    IID & 47.1{\scriptsize$\pm$0.1 } & 59.9{\scriptsize$\pm$0.1 } & 69.0{\scriptsize$\pm$0.3 } &  24.6{\scriptsize$\pm$0.1 } & 45.7{\scriptsize$\pm$0.4 } & {51.3{\scriptsize$\pm$0.4 }} & - & - & - \\ 
    DSDM & 45.0{\scriptsize$\pm$0.4 } & 66.5{\scriptsize$\pm$0.3 } & 75.8{\scriptsize$\pm$0.3 } &  19.5{\scriptsize$\pm$0.2 } & 46.2{\scriptsize$\pm$0.3 } & {54.0{\scriptsize$\pm$0.2 }} & - & - & - \\ 
    G-VBSM & - & 46.5{\scriptsize$\pm$0.7} & 54.3{\scriptsize$\pm$0.3} & 16.4{\scriptsize$\pm$0.7} & 38.7{\scriptsize$\pm$0.2} & 45.7{\scriptsize$\pm$0.4} & - & - & - \\
    \rowcolor{green!8} \textbf{\mymethod{} (Ours)} & \textbf{49.5{\scriptsize$\pm$0.3}} & \textbf{71.8{\scriptsize$\pm$0.3}} & \textbf{77.4{\scriptsize$\pm$0.3}} &  \textbf{34.4{\scriptsize$\pm$0.5}} & \textbf{48.7{\scriptsize$\pm$0.3}} &  \textbf{54.7{\scriptsize$\pm$0.2}} &  \textbf{18.2{\scriptsize$\pm$0.5}} & \textbf{26.8{\scriptsize$\pm$0.6}} & \textbf{29.6{\scriptsize$\pm$0.5}} \\
    \midrule
     Whole Dataset & \multicolumn{3}{c|}{84.8{\scriptsize$\pm$0.1 }}& \multicolumn{3}{c|}{56.2{\scriptsize$\pm$0.3 }}& \multicolumn{3}{c}{37.6{\scriptsize$\pm$0.4 }} \\ 
    \bottomrule
    \end{tabular}}
    }
    \vspace{-5pt}
\end{table*}

\section{Experiments}

\subsection{Setup}\label{sec:setup}

\noindent \textbf{Baseline methods.} We compared \mymethod{} with several representative approaches in dataset distillation and coreset selection. These include gradient-matching methods such as DC~\citep{DC}, DCC~\citep{DCC}, DSA and DSAC~\citep{DSA}. Kernel-based methods like KIP~\citep{KIP} and FrePo~\citep{FRePo} were also included. Distribution-matching methods like CAFE~\citep{CAFE}, DM~\citep{DM}, IDM~\citep{IDM}, M3D~\citep{M3D}, IID~\citep{IID}, and DSDM~\citep{DSDM} were part of the evaluation. We also included trajectory-matching methods such as MTT~\citep{MTT}, FTD~\citep{FTD}, ATT~\citep{ATT}, and TESLA~\citep{TESLA}. \textit{State-of-the-art} methods like DATM~\citep{DATM}, G-VBSM~\citep{G-VBSM}, and RDED~\citep{RDED} were also considered in our comparisons. Additionally, we benchmarked our method against classical coreset selection techniques, including random selection, Herding~\citep{herding}, and Forgetting~\citep{forgetting}.

\noindent \textbf{Datasets and Networks}. Our evaluations were conducted on widely-used datasets, including CIFAR-10 and CIFAR-100~\citep{krizhevsky2009learning} with resolution of 32×32, Tiny ImageNet~\citep{le2015tiny} with resolution of 64×64, and ImageNet subsets with resolution of 128×128, \emph{i.e.}, ImageNette, ImageWoof, ImageFruit, ImageMeow, ImageSquawk, and ImageYellow~\citep{imagenet_subset}. Following prior studies~\citep{DATM,drupi}, we used networks with instance normalization as the default setting. Specifically, dataset distillation is performed with a 3-layer ConvNet for CIFAR-10/100, a 4-layer ConvNet for Tiny ImageNet, and a 5-layer ConvNet for ImageNet subsets. All experiments were conducted with 10 evaluations for fairness, primarily using a single NVIDIA 4090 GPU.

\begin{table*}[tb!]
\centering
\caption{Results on ImageNet subsets (resolution of 128×128) when employing \mymethod{} across different IPCs. }
\vspace{-8pt}
\setlength{\tabcolsep}{2.5mm} 
\resizebox{0.95\textwidth}{!}{ 
\begin{tabular}{c|cc|cc|cc|cc|cc|cc}
\toprule
 Dataset &  \multicolumn{2}{c|}{ImageNette}  &  \multicolumn{2}{c|}{ImageWoof}  &  \multicolumn{2}{c|}{ImageFruit} &  \multicolumn{2}{c|}{ImageMeow} &  \multicolumn{2}{c|}{ImageSquawk} &  \multicolumn{2}{c}{ImageYellow} \\
IPC    & 1 & 10 & 1 & 10 & 1 & 10 & 1 & 10 & 1 & 10 & 1 & 10 \\
Ratio (\%)     & 0.105 & 1.050 & 0.110 &  1.100 & 0.077 & 0.77  & 0.077 & 0.77  & 0.077 & 0.77 & 0.077 & 0.77 \\
\midrule
Random     & 23.5{\scriptsize$\pm$4.8} & 47.7{\scriptsize$\pm$2.4} & 14.2{\scriptsize$\pm$0.9} & 27.0{\scriptsize$\pm$1.9} & 13.2{\scriptsize$\pm$0.8} & 21.4{\scriptsize$\pm$1.2} & 13.8{\scriptsize$\pm$0.6} & 29.0{\scriptsize$\pm$1.1} & 21.8{\scriptsize$\pm$0.5} & 40.2{\scriptsize$\pm$0.4} & 20.4{\scriptsize$\pm$0.6} & 37.4{\scriptsize$\pm$0.5} \\
MTT     & 47.7{\scriptsize$\pm$0.9} & 63.0{\scriptsize$\pm$1.3} & \textbf{28.6{\scriptsize$\pm$0.8}} & 35.8{\scriptsize$\pm$1.8} & 26.6{\scriptsize$\pm$0.8} & 40.3{\scriptsize$\pm$1.3} & 30.7{\scriptsize$\pm$1.6} & 40.4{\scriptsize$\pm$2.2} & 39.4{\scriptsize$\pm$1.5} & 52.3{\scriptsize$\pm$1.0} & 45.2{\scriptsize$\pm$0.8} & 60.0{\scriptsize$\pm$1.5} \\

DM     & 32.8{\scriptsize$\pm$0.5} & 58.1{\scriptsize$\pm$0.3} & 21.1{\scriptsize$\pm$1.2} & 31.4{\scriptsize$\pm$0.5} & - & - & - & - & 31.2{\scriptsize$\pm$0.7} & 50.4{\scriptsize$\pm$1.2} & - & - \\
RDED     & 33.8{\scriptsize$\pm$0.8} & 63.2{\scriptsize$\pm$0.7} & 18.5{\scriptsize$\pm$0.9} & 40.6{\scriptsize$\pm$2.0} & - & - & - & - & - & - & - & - \\
\rowcolor{green!8}\textbf{\mymethod{} (Ours)} & \textbf{53.4{\scriptsize$\pm$1.6}} & \textbf{77.6{\scriptsize$\pm$1.0}} & 27.2{\scriptsize$\pm$1.1} & \textbf{48.4{\scriptsize$\pm$1.3}} & \textbf{29.2{\scriptsize$\pm$0.7}} & \textbf{44.8{\scriptsize$\pm$1.5}} & \textbf{34.6{\scriptsize$\pm$0.7}} & \textbf{58.2{\scriptsize$\pm$1.2}} & \textbf{41.6{\scriptsize$\pm$1.2}} & \textbf{72.8{\scriptsize$\pm$0.9}} & \textbf{46.6{\scriptsize$\pm$1.5}} & \textbf{74.2{\scriptsize$\pm$1.4}} \\
\midrule
Whole Dataset &  \multicolumn{2}{c|}{87.4$\pm$1.0}  &  \multicolumn{2}{c|}{67.0$\pm$1.3}  &  \multicolumn{2}{c|}{63.9$\pm$2.0} &  \multicolumn{2}{c|}{66.7$\pm$1.1} &  \multicolumn{2}{c|}{87.5$\pm$0.3} &  \multicolumn{2}{c}{84.4$\pm$0.6} \\
\bottomrule
\end{tabular}
}
\label{tab:res_imagenetsubsets}
\vspace{-10pt}
\end{table*}

\noindent \textbf{Other Settings.} Following prior works, we implemented differential augmentation~\citep{DSA,CAFE} and applied multi-formation parameterization with a scale factor of $\rho=2$ for images, as in~\citep{IDC,IDM}. We employed AdamW as our optimizer. In our setup, we set the number of sampled frequency arguments to 1024. The number of mixture Gaussian components in the sampling network is set to the number of frequency arguments divided by 16, balancing the sampling network diversity and computational efficiency. Further details are provided in the supplementary material.

\subsection{Main Results}
We verified the effectiveness of \mymethod{} on various benchmark datasets of different image-per-class (IPC) settings\footnote{We provide further results on continual learning, neural architecture search, and larger IPC datasets in the supplementary material.}.

\noindent \textbf{CIFAR-10/100 and Tiny ImageNet.} As shown in Table~\ref{tab:dd_main_exp}, \mymethod{} outperforms all state-of-the-art (SOTA) baselines. Specifically, it surpasses distribution matching methods using traditional metrics like MSE and MMD, achieving improvements of 23.5\% and 23.0\% on CIFAR-10 and CIFAR-100 with 1 IPC compared to DM~\citep{DM}. Additionally, \mymethod{} maintains SOTA performance even against computationally intensive methods like MTT~\citep{MTT}. Results for larger IPC settings and comparisons with other SOTA methods like DATM~\citep{DATM} are in the supplementary material.

\noindent \textbf{Higher-resolution Datasets.} We also evaluated \mymethod{} on larger datasets, specifically the ImageNet subsets. As shown in Table~\ref{tab:res_imagenetsubsets}, \mymethod{} demonstrates strong performance across these challenging benchmarks. In 10 IPC setting, our method achieves substantial improvements of 20.5\% on ImageSquawk, compared to the baseline MTT~\citep{MTT}. Remarkably, \mymethod{} exhibits robust performance under relatively small IPC. For instance, compared to RDED~\citep{RDED}, \mymethod{} yields a significant improvement of 19.6\% on ImageNette.

\noindent \textbf{Computational Efficiency Evaluation.} \label{sec:efficiency}We tested the training speed and GPU memory of our \mymethod{} compared with \textit{strong} baseline methods on different datasets. As conventional recognition, trajectory matching based methods usually achieve better results than distribution matching in practice~\citep{MTT,TESLA,FTD}. However, both superior training efficiency and GPU memory efficiency are observed in \mymethod{} across all benchmark datasets, while achieving better results. Specifically, we measured the average training time over 1000 distillation iterations for each method, as summarized in Table~\ref{tab:speed_memory}. For CIFAR-100 at IPC 50, \mymethod{} achieves nearly 30$\times$ faster speeds compared to TESLA~\citep{TESLA} without the sampling network, and maintains over 20$\times$ faster speeds with the sampling network included. Moreover, we conducted a comprehensive analysis of computational efficiency, where GPU memory is expressed as the peak memory usage during 1000 iterations of training, as shown in Table~\ref{tab:speed_memory}. While most existing methods encounter out of memory (OOM) issues at IPC = 50, our method requires only about 1.9GB GPU memory on CIFAR-100. This further demonstrates the exceptional scalability of our approach under high IPC conditions. Further results on CIFAR-10 are provided in the supplementary material.

\begin{table}[tb!]
\centering
\caption{Training speed (s/iter) and peak GPU memory (GB) comparison on a single NVIDIA A100 80G. \textcolor{americanrose}{OOM} marks out-of-memory cases. `Reduction' shows \mymethod{}’s speed and memory improvements over the best-performing baseline in the table.}
\label{tab:speed_memory}
\vspace{-10pt}
\setlength\tabcolsep{2mm}
\renewcommand{\arraystretch}{1.2}
\resizebox{0.49\textwidth}{!}{
\begin{tabular}{@{}c|cc|cc|cc|cc@{}}
\toprule
Resource &\multicolumn{4}{c|}{Speed (s/iter)}&\multicolumn{4}{c}{GPU Memory (GB)}\\ \midrule
Dataset & \multicolumn{2}{c|}{CIFAR-100} & \multicolumn{2}{c|}{Tiny ImageNet} 
& \multicolumn{2}{c|}{CIFAR-100} & \multicolumn{2}{c}{Tiny ImageNet} \\
IPC & 10 & 50 & 10 & 50 
& 10 & 50 & 10 & 50 \\ \midrule

MTT                  & 1.92 & \textcolor{americanrose}{OOM}    & \textcolor{americanrose}{OOM}    & \textcolor{americanrose}{OOM}    & 61.6& \textcolor{americanrose}{OOM}    & \textcolor{americanrose}{OOM}  & \textcolor{americanrose}{OOM}  \\
FTD                  & 1.68 & \textcolor{americanrose}{OOM}    & \textcolor{americanrose}{OOM}    & \textcolor{americanrose}{OOM}    & 61.4& \textcolor{americanrose}{OOM}    & \textcolor{americanrose}{OOM}  & \textcolor{americanrose}{OOM}  \\
TESLA                & 5.71 & 28.24 & 42.01 & \textcolor{americanrose}{OOM}    & 10.3& 44.2& 69.6 & \textcolor{americanrose}{OOM}  \\
DATM                 & \textcolor{americanrose}{OOM}   & \textcolor{americanrose}{OOM}    & \textcolor{americanrose}{OOM}    & \textcolor{americanrose}{OOM}    & \textcolor{americanrose}{OOM}  & \textcolor{americanrose}{OOM}  & \textcolor{americanrose}{OOM}  & \textcolor{americanrose}{OOM}  \\ \midrule
\mymethod{} w/o $\psi$ & 0.73 & 0.96  & 2.40  & 5.67  & 1.4  & 1.9& 6.4& 8.4\\ 
\rowcolor{green!8} Reduction & \textbf{2.3$\times$} & \textbf{29.4$\times$} & \textbf{17.5$\times$} & \textbf{-} & \textbf{7.4$\times$} & \textbf{23.3$\times$}& \textbf{10.9$\times$}& \textbf{-} \\  \midrule
\mymethod{}       & 1.33 & 1.36  & 3.27  & 7.22  & 1.6  & 2.0& 6.5& 8.7\\ 
\rowcolor{green!8} Reduction& \textbf{1.3$\times$} & \textbf{20.8$\times$} & \textbf{12.8$\times$} & \textbf{-} & \textbf{6.4$\times$} & \textbf{22.1$\times$}& \textbf{10.7$\times$}& \textbf{-} \\  
\bottomrule
\end{tabular}
}
\vspace{-5pt}
\end{table}

\noindent \textbf{Cross-Architecture Generalization.} We evaluated the cross-architecture generalization capability of our method by testing its performance on various network architectures, including AlexNet~\citep{krizhevsky2009learning}, VGG-11~\citep{VGG}, and ResNet-18~\citep{resnet}. In this evaluation, synthetic data were condensed using a 3-layer ConvNet, and each method was subsequently tested across different architectures to assess robustness and adaptability. Tables~\ref{tab:cross_arch} summarize the results on CIFAR-10 with 10 and 50 IPC settings, respectively. In both cases, \mymethod{} consistently outperformed other methods across all architectures, demonstrating its strong ability to generalize effectively even when trained on a different architecture.  Results on other backbone networks beyond ConvNet are provided  in the supplementary material.
\begin{table}[htbp]
    \vspace{-5pt}
    \caption{Cross-architecture generalization performance (\%) on CIFAR-10. The synthetic data is condensed using ConvNet, and each method is evaluated on different architectures.}
    \vspace{-6pt}
    \label{tab:cross_arch}
    \centering
    \setlength\tabcolsep{3mm} 
    \renewcommand{\arraystretch}{1.15} 
    \resizebox{0.49\textwidth}{!}{
    \begin{tabular}{cc|cccc}
    \toprule
    IPC  & Method & ConvNet & AlexNet & VGG & ResNet \\ \midrule
     \multirow{4}{*}{10}& DSA & 52.1{\scriptsize$\pm$0.4} & 35.9{\scriptsize$\pm$1.3} & 43.2{\scriptsize$\pm$0.5} & 35.9{\scriptsize$\pm$1.3} \\
     & MTT & 64.3{\scriptsize$\pm$0.7} & 34.2{\scriptsize$\pm$2.6} & 50.3{\scriptsize$\pm$0.8} & 34.2{\scriptsize$\pm$2.6} \\
     & KIP & 47.6{\scriptsize$\pm$0.9} & 24.4{\scriptsize$\pm$3.9} & 42.1{\scriptsize$\pm$0.4} & 24.4{\scriptsize$\pm$3.9} \\
      & \cellcolor{green!8} \textbf{\mymethod{}}& \cellcolor{green!8} \textbf{71.8{\scriptsize$\pm$0.3}} & \cellcolor{green!8} \textbf{67.9{\scriptsize$\pm$0.5}} & \cellcolor{green!8} \textbf{68.0{\scriptsize$\pm$0.3}} & \cellcolor{green!8} \textbf{67.7{\scriptsize$\pm$0.5}} \\ \midrule
      \multirow{3}{*}{50} & DSA & 59.9{\scriptsize$\pm$0.8} & 53.3{\scriptsize$\pm$0.7} & 51.0{\scriptsize$\pm$1.1} & 47.3{\scriptsize$\pm$1.0} \\
     & DM & 65.2{\scriptsize$\pm$0.4} & 61.3{\scriptsize$\pm$0.6} & 59.9{\scriptsize$\pm$0.8} & 57.0{\scriptsize$\pm$0.9} \\
      &  \cellcolor{green!8} \textbf{\mymethod{}}& \cellcolor{green!8} \textbf{77.4{\scriptsize$\pm$0.3}} & \cellcolor{green!8} \textbf{75.5{\scriptsize$\pm$0.3}} & \cellcolor{green!8} \textbf{75.5{\scriptsize$\pm$0.3}} & \cellcolor{green!8} \textbf{73.8{\scriptsize$\pm$0.2}} \\ \bottomrule
    \end{tabular}}
    \vspace{-15pt}
\end{table}


\subsection{Ablation Study}

\subsubsection{Effect of the Sampling Network}\label{sec:sampling_net}

To rigorously evaluate the impact of the sampling network, $\psi$, within the minmax paradigm of \mymethod{}, we conducted performance comparisons with and without this component. To ensure a controlled and fair assessment, no additional data curation techniques were applied (such as fine-tuning or soft label integration). As shown in Table~\ref{tab:sample-net}, employing the sampling network $\psi$ yields substantial improvements in synthetic data quality across various datasets. For example, integrating $\psi$ into our method provides a 3.2\% performance increase on CIFAR-10 at 50 IPC. Our method yields a 2.6\% performance increase on Tiny ImageNet at 1 IPC and 10.1\% at 10 IPC. Similar trends are observed across ImageNet subsets, including gains of 2.8\% on ImageMeow and 2.0\% on ImageSquawk. The strong performance benefits from sampling network $\psi$ emphasize the effectiveness of the minmax paradigm compared to straightforward CFD minimization.

\begin{table*}[tb!]
\centering
\vspace{-10pt}
\caption{Test Performance (\%) on CIFAR-10, CIFAR-100, Tiny ImageNet and ImageNet subsets with and without the sampling network $\psi$. We find that  sampling network $\psi$ significantly improves performance, even without additional data curation steps.}
\vspace{-6pt}
\setlength{\tabcolsep}{3mm}
\resizebox{0.95\textwidth}{!}{
\begin{tabular}{c|cc|cc|ccc|c|c|c|c}
\toprule
 Dataset& \multicolumn{2}{c|}{CIFAR-10} & \multicolumn{2}{c|}{CIFAR-100} & \multicolumn{3}{c|}{Tiny ImageNet}&ImageFruit&ImageMeow&ImageSquawk&ImageYellow\\
IPC & 10 & 50 & 10 & 50  &1 & 10 & 50 & 10 & 10 & 10 & 10\\
\midrule
\mymethod{} w/o $\psi$ & 65.6 & 74.2 & 45.9 & 53.7  &9.4 & 14.2& 22.0 &39.6 & 51.6 & 68.8 &67.6\\
\rowcolor{green!8}\textbf{\mymethod{}} & \textbf{68.9} & \textbf{77.4} & \textbf{48.7} & \textbf{54.4} & \textbf{12.0} & \textbf{24.3}& \textbf{26.5} &\textbf{41.4} & \textbf{54.4} & \textbf{70.8} & \textbf{69.2} \\
\bottomrule
\end{tabular}
}
\label{tab:sample-net}
\vspace{-5pt}
\end{table*}

\subsubsection{Impact of Amplitude and Phase Components}\label{sec:amplitude_phase}

\begin{figure}[htbp]
    \centering
    \vspace{-5pt}
    \includegraphics[width=0.99\linewidth]{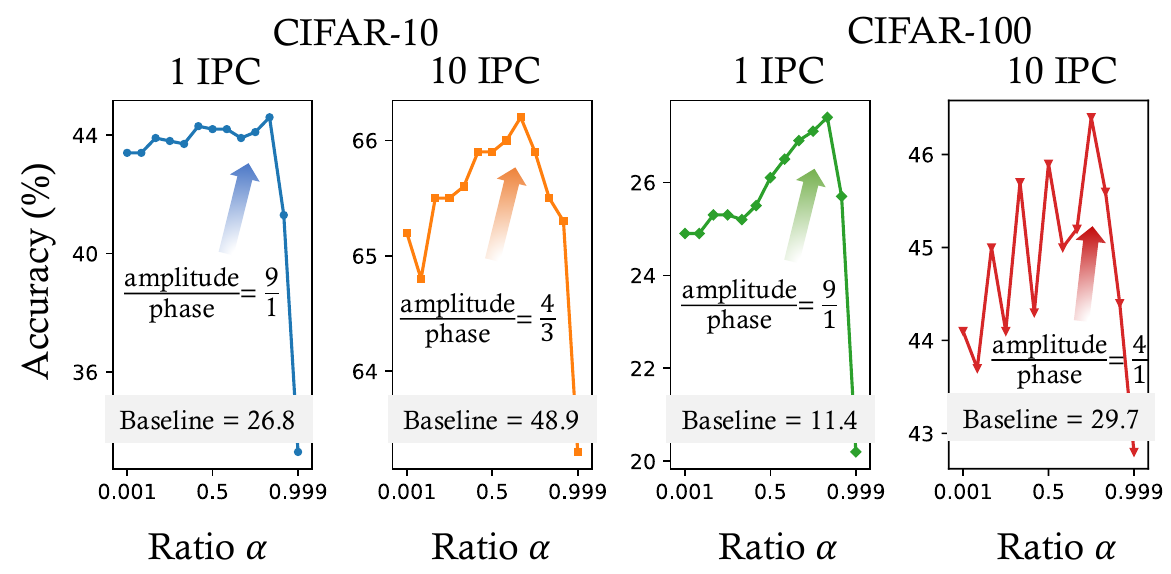}
    \vspace{-10pt}
    \caption{Impact of amplitude and phase components in the NCFD measure across various datasets and IPC settings. The figure illustrates the relationship between the amplitude-to-phase ratio $\alpha$ in Eq.~(\ref{eq:alpha_blend}). Results indicate that balancing amplitude (for diversity) and phase (for realism) information leads to improved performance. Baseline results were obtained using DM~\citep{DM}.}
    \label{fig:amplitude_phase}
    \vspace{-15pt}
\end{figure}

We examine individual contributions of amplitude and phase alignment within the NCFD measure. By selectively adjusting amplitude or phase alignment, controlled by the hyperparameter $\alpha$ that represents the ratio of amplitude to phase weight in the loss function, we find that both components are essential. To further evaluate the effect of $\alpha$ on performance, we conducted ablation studies on the CIFAR-10 and CIFAR-100 datasets. As noted in prior works~\cite{mandic2009complex, oppenheim1981importance}, the amplitude term primarily enhances the diversity of generated data, while the phase term contributes to realism by accurately capturing data centers. For example, as shown in Figure~\ref{fig:amplitude_phase}, on CIFAR-10 with 10 IPC, when the amplitude information dominates the loss (\emph{e.g.}, $\alpha=0.999$), the test accuracy decreases about 3\% compared to our best results. Conversely, when the phase information dominates (\emph{e.g.}, $\alpha=0.001$), the test accuracy decreases by about 1\%. Results demonstrate that a balanced integration of both components yields the highest accuracy.

\subsubsection{Effect of the Number of Sampled  Frequency \\ Arguments in NCFD}\label{sec:n_freq}

\begin{figure}[htbp]
    \centering
    \vspace{-10pt}
    \includegraphics[width=0.75\linewidth]{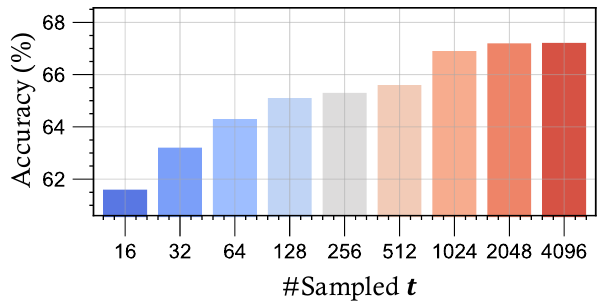}
    \vspace{-8pt}
    \caption{Impact of sampled frequency count in NCFD on accuracy across datasets and IPC. Increasing frequencies improves accuracy up to a threshold, beyond which gains diminish.}
    \label{fig:n_freq}
    \vspace{-15pt}
\end{figure}

To assess the impact of the number of sampled frequency arguments, $\bm{t}$, generated by the sampling network $\psi$, we varied the sample count and measured the corresponding performance. As illustrated in Figure~\ref{fig:n_freq}, increasing the number of sampled arguments initially enhances the quality of synthetic data by facilitating finer distributional alignment. For example, accuracy on CIFAR-10 at 10 IPC improves from 62\% with 16 sampled frequency arguments to approximately 67
\% with 1024, indicating a positive correlation between the sampled number and accuracy. However, beyond 1024 arguments, performance gains plateau, with accuracy stabilizing around 67-68\% even as the sampling number increases to 4096. This trend suggests that a moderate number achieves an optimal balance between computational efficiency and accuracy. We observed that  additional cost remains \textit{minimal} as the number of sampled arguments increases, underscoring \mymethod{}’s ability to produce high-quality synthetic data with low computational cost.

\section{Discussion}
\subsection{Training stability of NCFD}
\begin{figure}[htbp]
    \centering
\includegraphics[width=0.75\linewidth]{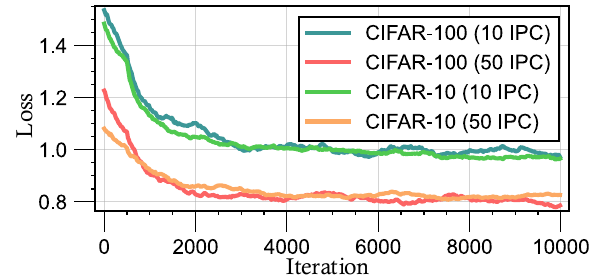}
    \vspace{-7pt}
    \caption{Training dynamics of the minmax optimization process across different datasets and various IPC settings.}
    \vspace{-2pt}
    \label{fig:stability}
\end{figure}

The training stability of our minmax paradigm is crucial to its effectiveness. Unlike traditional discrepancy measures, \mymethod{} operates within the complex plane to conduct minmax optimization. While instability is a common issue in minmax adversarial optimization, as seen in generative adversarial networks~\cite{salimans2016improved,arjovsky2017towards,radford2015unsupervised}, \mymethod{} consistently maintains stable optimization throughout training, as illustrated in Figure~\ref{fig:stability}. This stability is further supported by theoretical guarantees of weak convergence in Theorem~\ref{thm:convergence}, demonstrating the robustness of the CF-based discrepancy under diverse conditions and contributing to \mymethod{}'s reliable convergence across datasets.

\subsection{Correlation between CFD and MMD}\label{sec:correlation}
To better understand \mymethod{}, we examine the relationship between the Characteristic Function Discrepancy (CFD) and Maximum Mean Discrepancy (MMD). 

\noindent \textbf{CF as Well-Behaved Kernels in the MMD Metric.} The CF discrepancy term $\int_{\bm{t}} \sqrt{\textrm{Chf}(\bm{t};f)} dF_{\mathcal{T}}(\bm{t})$ in our loss can be viewed as a well-behaved kernel in MMD, specifically as a \textit{Characteristic Kernel}~\citep{sriperumbudur2010hilbert}. Unlike MMD, which relies on fixed kernels, \mymethod{} adaptively learns $F_{\mathcal{T}}(\bm{t})$, enabling flexible kernel selection for optimal distribution alignment. Furthermore, mixtures of Gaussian distributions within the CF framework produce well-defined characteristic kernels. When MMD employs a characteristic kernel of the form $\int_{\bm{t}} e^{-j \langle \bm{t}, \bm{x} - \bm{\tilde{x}} \rangle} dF_{\mathcal{T}}(\bm{t})$, it aligns with the structure of CFD, demonstrating that \textit{MMD is a special case of CFD} when only specific moments are matched. This insight also explains the minimal memory overhead observed as IPC grows, highlighting the efficiency of our approach.

\noindent \textbf{Computational Advantage of CFD over MMD.} In contrast to MMD, which requires \textit{quadratic} time in the number of samples for approximate computation, CFD operates in \textit{linear} time relative to the sampling number of frequency arguments, which aligns results in~\citep{CFGAN}. This efficiency makes CFD substantially faster and more scalable than MMD, offering a particular advantage for large-scale datasets.

%% file: sec/5_conclusion.tex
\section{Conclusion}
In this work, we redefined distribution matching for dataset distillation as a minmax optimization problem and introduced Neural Characteristic Function Discrepancy (NCFD), a novel and theoretically grounded metric designed to maximize the separability between real and synthetic data. Leveraging the Characteristic Function (CF), our method dynamically adjusts NCFD to align both phase and amplitude information in the complex plane, achieving a balance between realism and diversity. Extensive experiments demonstrated the computational efficiency of our approach, which achieves state-of-the-art performance with minimal computational overhead, showcasing its scalability and practicality for large-scale applications.